\documentclass[letterpaper]{article} 
\usepackage{aaai25}  
\usepackage{times}  
\usepackage{helvet}  
\usepackage{courier}  
\usepackage[hyphens]{url}  
\usepackage{graphicx} 
\usepackage{natbib}  
\usepackage{caption} 
\usepackage{amsmath}
\usepackage{amssymb}
\usepackage{tikz}
\usepackage{booktabs}  
\usepackage{graphicx}
\usepackage{amsmath}
\usepackage{booktabs}
\usepackage{multirow}
\usepackage{makecell}
\usepackage{algorithm}
\usepackage{algorithmic}
\usepackage{tabularx}  
\usepackage{cleveref}

\usepackage{newfloat}
\usepackage{listings}
\DeclareCaptionStyle{ruled}{labelfont=normalfont,labelsep=colon,strut=off} 
\floatstyle{ruled}
\newfloat{listing}{tb}{lst}{}
\floatname{listing}{Listing}
\nocopyright 

\title{Estimating Peer Direct and Indirect Effects in Observational Network Data}

\author {
    Xiaojing Du\textsuperscript{\rm 1},
    Jiuyong Li\textsuperscript{\rm 1},
    Debo Cheng\textsuperscript{\rm 1},
    Lin Liu\textsuperscript{\rm 1},
    Wentao Gao\textsuperscript{\rm 1},
    Xiongren Chen\textsuperscript{\rm 1}
}
\affiliations {
    \textsuperscript{\rm 1}University of South Australia\\
    \{xiaojing.du, wentao.gao, xiongren.chen\}@mymail.unisa.edu.au,
    \{jiuyong.li, debo.cheng, liu.lin\}@unisa.edu.au
}

\begin{document}
\maketitle

\begin{abstract}
Estimating causal effects is crucial for decision-makers in many applications, but it is particularly challenging with observational network data due to peer interactions. Many algorithms have been proposed to estimate causal effects involving network data, particularly peer effects, but they often overlook the variety of peer effects. To address this issue, we propose a general setting which considers both peer direct effects and peer indirect effects, and the effect of an individual's own treatment, and provide identification conditions of these causal effects and proofs. To estimate these causal effects, we utilize attention mechanisms to distinguish the influences of different neighbors and explore high-order neighbor effects through multi-layer graph neural networks (GNNs). Additionally, to control the dependency between node features and representations, we incorporate the Hilbert-Schmidt Independence Criterion (HSIC) into the GNN, fully utilizing the structural information of the graph, to enhance the robustness and accuracy of the model. Extensive experiments on two semi-synthetic datasets confirm the effectiveness of our approach. Our theoretical findings have the potential to improve intervention strategies in networked systems, with applications in areas such as social networks and epidemiology.

\end{abstract}

%

\section{Introduction}
  
Causal effect estimation is an important area of study, with the focus on determining cause-and-effect relationships between variables~\cite{imbens2010rubin,pearl2009causality}. It is challenging to achieve accurate causal effect estimation using observational data due to the presence of confounders that affect both the treatment and the outcome. When the data is collected from sources such as social networks, communication networks, or biological networks, causal effect estimation becomes even more challenging since the data is inherently networked, meaning that units (e.g., individuals, nodes) are interconnected, and their outcomes potentially being influenced by both the treatments and outcomes of their neighbors~\cite{sinclair2012detecting}. This implies that traditional causal inference methods are not applicable any more since they may not adequately account for peer effects or network dependencies~\cite{yao2021survey}. Therefore, estimating causal effects on network data necessitates specialized techniques that can handle these interdependencies, ensuring that the estimated effects are not biased by the network structure itself.

 \begin{figure}[t]
	\centering
	\includegraphics[scale=0.26]{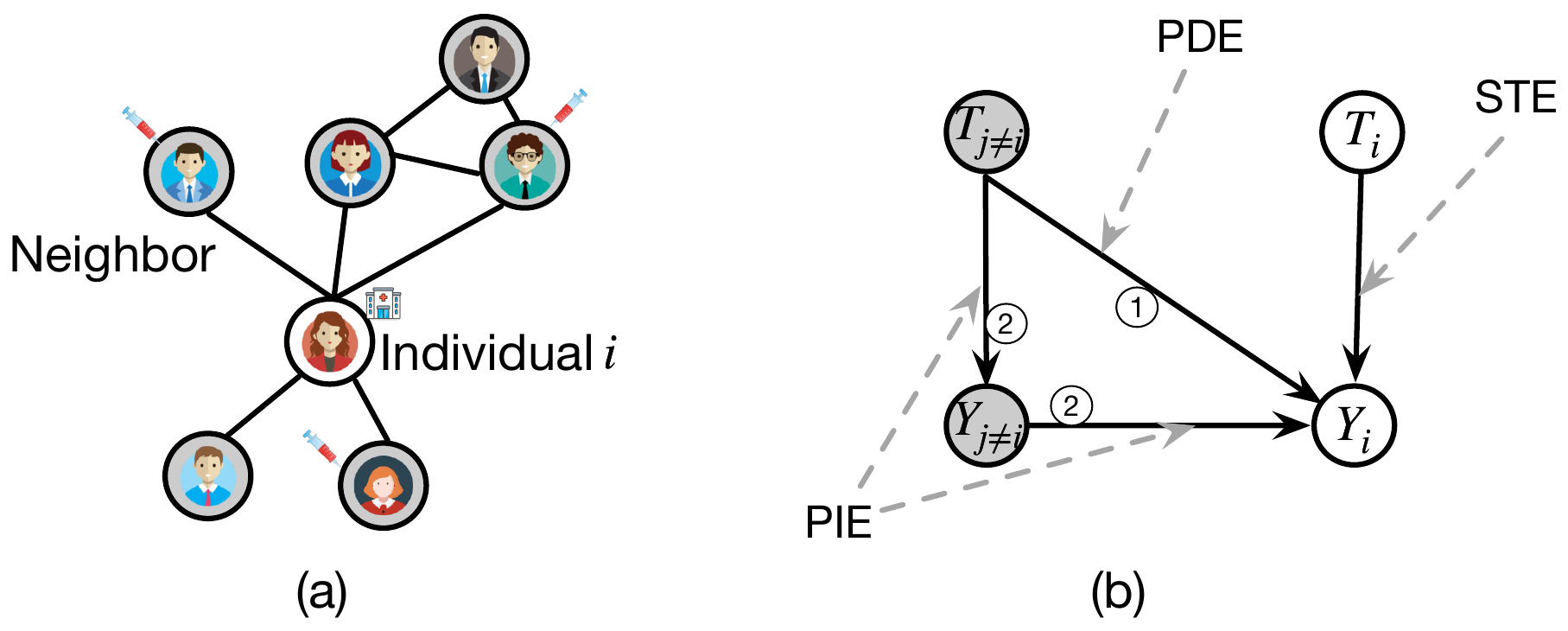}
	\caption{(a) A graph to showing the relationship between individual \( i \) and their neighbors in network data. (b) A causal graph illustrating  peer direct effects (PDE), peer indirect effects (PIE), and self-treatment effects (STE). In the diagram, $T$ and $Y$ indicate the intervention (e.g., vaccination status and infection condition), the subscript \( i \) indicates an individual and \({j \neq i} \) indicates their neighbours.}
	\label{fig:1}
\end{figure}

Effects in network data can be classified into peer effects (PE, an individual's treatment affects another's outcome) and self-treatment effects (STE, an individual's treatment affects their own outcome). As illustrated in Fig.~\ref{fig:1}(a), in the context of an infectious disease, an individual's infection status depends on both their own and their neighbors' vaccination statuses. Peer effects can be direct, where the vaccination of an individual's neighbour reduces the risk of the individual's infection (\textit{peer direct effect}, PDE, as indicated by the edge marked with \normalsize{\textcircled{\scriptsize{1}}} in Fig.~\ref{fig:1}(b)), or indirect, where the vaccination of an individual's neighbour affects the neighbor's own infection status, which in turn (indirectly) affects the individual's infection status (\textit{peer indirect effect}, PIE, as indicated by the path marked with \normalsize{\textcircled{\scriptsize{2}}} in Fig.~\ref{fig:1}(b)). Understanding these mechanisms is crucial for optimizing vaccination strategies. Peer direct effect denotes the effectiveness of a vaccine on an individual by vaccinating their neighbours regardless of their neighbours' infectious conditions. Peer indirect effect indicates the effectiveness of a vaccine on an individual by improving their neighbours infectious conditions through vaccination.

\begin{table*}[t]
\centering
\footnotesize 
\setlength{\tabcolsep}{6pt} 
\begin{tabularx}{\textwidth}{X l p{4cm}} 
\toprule
\textbf{Method} & \textbf{Type of Relationship} & \textbf{Causal Effects Considered} \\ \midrule
VanderWeele \cite{vanderweele2012components} & Unidirectional, one to one & PDE, PIE \\ 
Shpitser \cite{Shpitser2017ModelingIV} & Bidirectional, one to one & PDE, PIE \\ 
NetEst \cite{jiang2022estimating}, Cai \cite{cai2023generalization}, \newline TNet \cite{chen2024doubly} & Many to one (at group level) & PE, STE \\ 
gDIS (ours) & Many to one (at group level) & PDE, PIE, STE \\ \bottomrule
\end{tabularx}
\caption{Comparison of problem settings across different methods. Methods are identified by the authors' names when the method name is not specified in the original text.}
\label{tab:comparison}
\end{table*}

Causal mediation analysis~\cite{mackinnon2007mediation} is usually used to distinguish between direct and indirect effects in independent and identically distributed (IID) data. VanderWeele et al.~\cite{vanderweele2012components} combined causal mediation analysis with statistical models to explore how vaccinating a one-year-old child might affect the child's mother, especially within a two-person household. Shpitser et al.~\cite{Shpitser2017ModelingIV} further advanced this approach by developing a symmetric treatment decomposition method based on chain graphs to address peer effects. This method enables the analysis of complex scenarios involving mutual influence, such as the interaction between a child and their mother. It allows the overall peer effect to be decomposed into specific components, facilitating the analysis of how interference effects are transmitted between individuals and helping to distinguish between direct and indirect effects.

Jiang et al.~\cite{jiang2022estimating} formalized the estimation of network causal effects as a multi-task learning problem and introduced a framework called Networked Causal Effects Estimation (NetEst). This method utilizes GNNs~\cite{hu2020gpt} to capture feature representations of both individual and their first-order neighbors. Cai et al.~\cite{cai2023generalization} derived generalization bounds for causal effect estimates from network data and proposed a weighted regression strategy based on joint propensity scores~\cite{lee2010improving} combined with representation learning. Chen et al.~\cite{chen2024doubly} integrated the targeted learning~\cite{van2011targeted} into neural network training, developing a causal effect estimator.


As shown in Table~\ref{tab:comparison}, although VanderWeele et al.~\cite{vanderweele2012components} and Shpitser et al.~\cite{Shpitser2017ModelingIV} considered both peer direct effects and indirect effects, their work primarily focused on scenarios with a propagation unit of two people only, which limits their practical applicability. Additionally, they did not account for STE, which are significant in networks and crucial for policy-making in interventions. While the recent work by Jiang et al.~\cite{jiang2022estimating}, Cai et al.~\cite{cai2023generalization}, and Chen et al.~\cite{chen2024doubly} considered group-level peer effects, they did not differentiate between the various types of peer effects. In many cases, particularly in infectious diseases, it is not enough to simply know that overall peer effects. It is important to distinguish between PDE and PIE, as each has different implications.


To address these limitations, we consider a more general setting for observational network data, where we estimate PDE and PIE at group level, as well as STE within networks. We apply the principles of mediation analysis~\cite{pearl2014interpretation} and the backdoor criterion~\cite{pearl2009causality} to analyze these effects, and provide their identification conditions and the proofs. Based on the theoretical analysis, we develop gDIS for \underline{\textbf{g}}roup-level P\underline{\textbf{D}}E and P\underline{\textbf{I}}E, and \underline{\textbf{S}}TE estimation with network data.
To effectively capture the complexity of network effects, gDIS employs a multi-layer GNN to focus on high-order neighbor interference and leverages the potential of attention mechanisms~\cite{niu2021review} to account for the varying influence weights of different neighbors on each individual. Furthermore, to better control the dependency relationships between node features, we integrate the Hilbert-Schmidt Independence Criterion (HSIC)~\cite{ahmad2021regularized} into the GNN, fully leveraging the structural information of computational graphs.

Our main innovations are summarized as follows:

\begin{itemize}
\item We propose to decompose peer effects into direct and indirect components at the group level in observational network data, and provide both the theoretical analyses and the corresponding proofs for the identifiability of these effects    .

\item  We develop a novel method gDIS for estimating the two different types of peer effects at group level and the self-treatment effect, using network data.

\item We validated the effectiveness and robustness of the gDIS on semi-synthetic datasets, showing that it maintains great performance even in complex network data.
\end{itemize}

\section{Preliminary}
\label{section:preliminary}

This section provides the notations and problem setting used throughout the paper.

\subsection{Notations and Problem Setting}

Throughout the paper, uppercase letters (e.g., $T_i$, $W_{t_i}$) denote variables, while lowercase letters (e.g., $t_i$, $w_{t_i}$) represent their values. Bold uppercase letters (e.g., $\mathbf{W}_x$, $\mathbf{X}$, $\mathbf{W}_y$, $\mathbf{W}_{x_i}$) indicate sets of variables, vectors or matrices, and bold lowercase letters (e.g., $\mathbf{w}_x$, $\mathbf{x}$, $\mathbf{w}_y$, $\mathbf{w}_{x_i}$) represent their corresponding values.

A network can be represented as a pair $(\mathbf{V}, \mathbf{E})$, where $\mathbf{V} = \{V_1, \ldots, V_m\}$ is the set of nodes, and $\mathbf{E} \subseteq \mathbf{V} \times \mathbf{V}$ is the set of edges between the nodes. A node $V_i \in \mathbf{V}$ has a set of features $\mathbf{X}_i = \{X_{i1}, \ldots, X_{ik}\}$, a treatment $T_i$, and an outcome $Y_i$ associated with it. We assume $T_i \in \{0, 1\}$ is a binary treatment, representing, e.g., whether node $V_i$ receives a vaccination or not, and $Y_i$ is a continuous variable, indicating, e.g., the level of immunity of $V_i$. For simplicity of presentation, we will use the index of a node, e.g., $i$, to represent the corresponding node $V_i$ when there is no ambiguity in the rest of the paper.

\begin{figure}[t]
	\centering
	\includegraphics[scale=0.35]{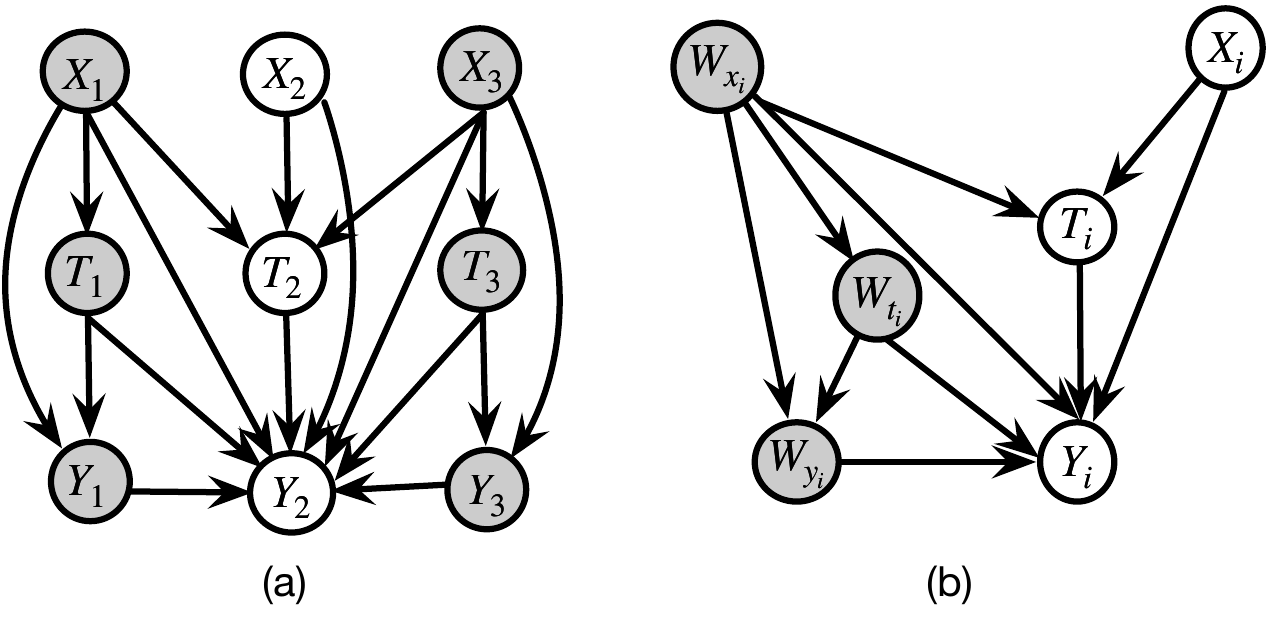}
	\caption{(a) An illustration of the causal relationships considered in our work, for node 2, which has node 1 and 3 as neighbours in the network. The features, treatment, and outcome of node \(i\) are represented by \(X_i\), \(T_i\), and \(Y_i\), respectively. (b) The summary causal graph where \(W_{x_i}\), \(W_{t_i}\), and \(W_{y_i}\) represent the aggregated features, treatments, and outcomes of node \(i\)'s neighbors.}
 
	\label{fig:2}
\end{figure}

Considering an individual or node \(i\), its outcome \(Y_i\) is affected by several factors: its own features \(\mathbf{X}_i\), its treatment \(T_i\), the features of its neighbors \(\{\mathbf{X}_j\}_{j \in \mathcal{N}_i}\), the treatments administered to those neighbors \(\{T_j\}_{j \in \mathcal{N}_i}\), and the outcomes observed in those neighbors \(\{Y_j\}_{j \in \mathcal{N}_i}\). Here, \(\mathcal{N}_i\) represents the set of neighbors of unit \(i\), as illustrated in Fig.~\ref{fig:2}(a).

We use Kullback-Leibler (KL) divergence~\cite{belov2011distributions} to compute the difference between the probability distributions \(P_i\) and \(P_j\) of node \(i\) and its neighboring node \(j\), as given by Eq~\ref{eq:kl} below:
\begin{equation}
D_{\text{KL}}(P_i \parallel P_j) = \sum_{k} P_i(k) \log ( \frac{P_i(k)}{P_j(k)})
\label{eq:kl}
\end{equation}
where \(k\) is the dimension of the feature vector of a node.

Based on the KL divergence, we calculate the influence weight of neighbor \(j\) on individual \(i\), \(w_{ij}\) as follows:
\begin{equation}
w_{ij} = \frac{1}{1 + D_{\text{KL}}(P_i \parallel P_j)}
\label{eq:wij}
\end{equation}
The higher the weight \(w_{ij}\), the greater the influence of neighbor \(j\) on individual \(i\).

Neighbor treatment exposure $W_{t_i}$, neighbor contagion exposure $W_{y_i}$, and neighbor feature $\mathbf{W}_{x_i}$ are calculated as:
$W_{t_i} = \sum_{j \in \mathcal{N}_i} w_{ij} T_j$, $W_{y_i} = \sum_{j \in \mathcal{N}_i} w_{ij} Y_j$, $\mathbf{W}_{x_i} = \sum_{j \in \mathcal{N}_i} w_{ij} \mathbf{X}_j$, respectively.

Our problem definition is provided as follows:

\textbf{Problem Definition.} Given the observational networked data of a network $(\textbf{V}, \textbf{E})$, including the data of the features, treatments and outcomes associated with the nodes of the network, our goal is for each node or individual $i$ to distinguish and obtain unbiased estimation of the \textit{peer direct effects} (PDE) (i.e., the causal effect of $W_{t_i}$ on $Y_i$ via $W_{t_i} \rightarrow Y_i$), \textit{peer indirect effects} (PIE) (i.e., the causal effect of $W_{t_i}$ on $Y_i$ via $W_{t_i} \rightarrow W_{y_i} \rightarrow Y_i$), and \textit{self-treatment effects} (STE) (i.e., the causal effect of $T_i$ on $Y_i$ via $T_i \rightarrow Y_i$) as shown in Fig.~\ref{fig:2}.
\label{problem-definition}

The formal definitions of group level PDE and PIE, and STE are based on the potential outcome framework~\cite{imbens2010rubin}. In the following, we first present the basic concepts of the framework, then give the definitions of PDE, PIE and STE.

For each unit \(i\) under two treatment conditions \(T_i\) (where \(T_i=0\) represents no treated unit and \(T_i=1\) represents treated unit), we define two potential outcomes: \(Y_i(1)\), the outcome if the unit receives treatment, and \(Y_i(0)\), the outcome if untreated.

The individual treatment effect (ITE) is defined as the difference between these potential outcomes:
\begin{equation}
\mathrm{ITE}_i = Y_i(1) - Y_i(0)
\label{eq:ITE}
\end{equation}
The average treatment effect (ATE) across the population is defined as the expected value of the ITE for all units:
\begin{equation}
\mathrm{ATE} = \mathbb{E}[Y_i(1) - Y_i(0)]
\label{eq:ATE}
\end{equation}

Under the potential outcome framework and following the definitions of PDE and PIE in~\cite{vanderweele2012components}, in our problem setting, we have: 
\begin{align}
\label{doPDE}
\mathrm{PDE} = \mathbb{E}\left[ Y_i(w_{t_i}', {W}_y(w_{t_i}')) - Y(w_{t_i}, {W}_y(w_{t_i}') \right]
\end{align}
\begin{align}
\label{doPIE}
\mathrm{PIE} = \mathbb{E}\left[ Y_i(w_{t_i},  {W}_y(w_{t_i}') - Y_i(w_{t_i}, {W}_y(w_{t_i})) \right]
\end{align}

The PDE defined in Eq.~\ref{doPDE} indicates the average of the difference between the potential outcomes of an individual when the aggregated treatment of its neighbours, i.e., its neighbor treatment exposure $W_{t_i}$, is changed from one level ($w_{t_i}$) to another ($w_{t_i}'$), while the aggregated neighbor potential outcome, i.e., its neighbor contagion exposure $W_{y_i}$, remains at the level as it would have been if the neighbours' aggregated treatment had been $w_{t_i}'$, denoted as $w_y(w_{t_i}')$. In the vaccination example, PDE indicates the average effect of changing the vaccination status of an individual's neighbors on the infectious status of the individual, assuming the neighbors' infectious status is kept at the level as after their vaccination status is changed.

The PIE defined in Eq.~\ref{doPIE} indicates the average of the difference between the potential outcomes of an individual when the aggregated treatment of its neighbours, i.e., its neighbor treatment exposure $W_{t_i}$, remains unchanged at ($w_{t_i}$), while the aggregated neighbor potential outcome, i.e., its neighbor contagion exposure $W_{y_i}$, is changed from what it would have been if the neighbours' aggregated treatment had been $w_{t_i}'$, denoted as  $w_y(w_{t_i}')$, to what it would have been if the neighbours' aggregated treatment had been $w_{t_i}$, i.e., $w_y(w_{t_i})$. In the vaccination example, PIE indicates that assuming the vaccination status of the neighbors of an individual remains unchanged, how the change of the infectious status of an individual's neighbors as a result of changing their vaccination status would affect the infectious status of the individual.

The definition of STE in our problem setting is similar to the definition of ATE, i.e., $\mathrm{STE} = \mathbb{E}\left[ Y_i(1) - Y_i(0) \right]$. In the vaccination example, STE indicates the average difference of potential outcomes of an individual if the individual had been vaccinated versus if they had not been vaccinated.


Estimating these effects deepens our understanding of interactions within complex networks and provides critical insights for informed policy decisions. 

\subsection{Assumptions}

In observational data, we only have the observed outcome, i.e., the factual outcome, but estimating causal effects requires the potential outcomes, i.e., both the factual and counterfactual outcomes. Therefore, the following assumptions are needed for estimating causal effects from observational network data~\cite{jiang2022estimating}.

In the following assumptions, \( Z_i = (W_{t_i}, W_{y_i}) \), where \( W_{t_i} \) and \( W_{y_i} \) represent the aggregated treatment and contagion exposures of node \( i \)'s neighbors, respectively. We define \( z_i \) as the corresponding value of \( Z_i \).

\textbf{Assumption 1} (Network Unconfoundedness): The potential outcome is independent of both individual treatment and neighborhood exposure, given the individual and neighbor features. i.e., \( Y_i(t_i, z_i) \perp\!\!\!\perp t_i, z_i \mid \mathbf{X}_i, \mathbf{W}_{x_i} \).

\textbf{Assumption 2} (Network Interference): If the potential outcome \( Y_i(t_i, z_i) \) for individual \(i\) depends on both \(i\)'s treatment \(T_i\) and \(Z_i\), network interference exists.

\textbf{Assumption 3} (Network Consistency): The potential outcome equals the observed outcome when a unit is exposed to the same treatment and neighborhood exposure. i.e., \( Y_i = Y_i(t_i, z_i) \) if unit \( i \) is subjected to \( t_i \) and \( z_i \).

\textbf{Assumption 4} (Network Overlap): Every treatment and neighborhood exposure pair \((T_i, Z_i)\) must have a positive probability of occurring. i.e., \( 0 < p(t_i, z_i \mid\mathbf{X}_i, \mathbf{W}_{x_i}) < 1 \).

\section{The Proposed gDIS Method}

\subsection{Identifiability of PDE, PIE and STE}

Establishing the identifiability of a causal effects is the prerequisite for estimating the causal effects from data. To study the identifiability of the causal effects (PDE, PIE and STE) with network data, we propose a novel causal graph, as shown in Fig.~\ref{fig:2}(b), to illustrate these causal effects. In this section, we provide the assumptions and conditions for identifying these causal effects from network data, along with the theoretical analysis and corresponding proofs.

To present our main theoretical results (Theorems 1 and 2 below), we make the following assumptions, which are commonly found in the causal inference literature and adapted to our problem setting, with causal relationships between variables illustrated in Fig.~\ref{fig:2}(b). For simplicity, we omit the subscript \( i \) from the names of the variables shown in Fig.~\ref{fig:2}(b) in the following discussion. 

\paragraph{Assumption 5} (Sequential Ignorability~\cite{imai2010identification}) 
There exists a set of observed covariates \(\mathbf{W}\) such that:

1\textsubscript{G}-1. \(\mathbf{W} \) and \( \mathbf{W}_t \) block all backdoor paths from \( \mathbf{W}_y \) to \( \mathbf{Y} \) that do not pass through \( \mathbf{W}_t \); 

1\textsubscript{G}-2. The set \( \mathbf{W} \) blocks all backdoor paths from \( \mathbf{W}_t \) to \( \mathbf{W}_y \) or \( \mathbf{Y} \), and no element in \( \mathbf{W} \) is a descendant of \( \mathbf{W}_t \).

A set \( \mathbf{W} \) that satisfies both conditions in Assumption 5 serves as an adjustment set for obtaining unbiased estimates of PDE and PIE from network data. We present our first theoretical finding below.

\paragraph{Theorem 1.} In the causal DAG represented in Fig~\ref{fig:2}(b), the set of neighbor features $\mathbf{W}_x$ satisfies both conditions 1\textsubscript{G}-1 and 1\textsubscript{G}-2 of Assumption 5.

\textit{Proof}: First, we prove that \(\{\mathbf{W}_x\}\) satisfies condition 1\textsubscript{G}-1 of Assumption 5. In the causal DAG shown in Fig.~\ref{fig:2}(b), all \( \mathbf{W}_t \)-avoiding backdoor paths from \(\mathbf{W}_y\) to \(\mathbf{Y}\) (\( \mathbf{W}_y \leftarrow \mathbf{W}_x \rightarrow \mathbf{Y}\), \( \mathbf{W}_y \leftarrow \mathbf{W}_x \rightarrow \mathbf{T} \leftarrow \mathbf{X} \rightarrow \mathbf{Y}\), and \( \mathbf{W}_y \leftarrow \mathbf{W}_x \rightarrow \mathbf{T} \rightarrow \mathbf{Y}\)) are blocked by the set \(\{\mathbf{W}_x, \mathbf{W}_t\}\), thus \(\{\mathbf{W}_x\}\) satisfies 1\textsubscript{G}-1. Next, we prove that \(\{\mathbf{W}_x\}\) satisfies condition 1\textsubscript{G}-2. The set \(\{\mathbf{W}_x\}\) blocks all backdoor paths from \( \mathbf{W}_t \) to \( \mathbf{W}_y \) (\( \mathbf{W}_t \leftarrow \mathbf{W}_x \rightarrow \mathbf{W}_y \), \( \mathbf{W}_t \leftarrow \mathbf{W}_x \rightarrow \mathbf{Y} \leftarrow \mathbf{W}_y\), \( \mathbf{W}_t \leftarrow \mathbf{W}_x \rightarrow \mathbf{T} \leftarrow \mathbf{Y} \leftarrow \mathbf{W}_y\), and \(\mathbf{W}_t \leftarrow \mathbf{W}_x \rightarrow \mathbf{T} \leftarrow \mathbf{X} \rightarrow \mathbf{Y} \leftarrow \mathbf{W}_y\)) and from \( \mathbf{W}_t \) to \( \mathbf{Y} \) (\( \mathbf{W}_t \leftarrow \mathbf{W}_x \rightarrow \mathbf{Y} \), \( \mathbf{W}_t \leftarrow \mathbf{W}_x \rightarrow \mathbf{W}_y \rightarrow \mathbf{Y} \), \( \mathbf{W}_t \leftarrow \mathbf{W}_x \rightarrow \mathbf{T} \rightarrow \mathbf{Y} \), and \(\mathbf{W}_t \leftarrow \mathbf{W}_x \rightarrow \mathbf{T} \leftarrow \mathbf{X} \rightarrow \mathbf{Y}\)). \(\mathbf{W}_x\) is not a descendant of \( \mathbf{W}_t \) in Fig.~\ref{fig:2}(b) and satisfies 1\textsubscript{G}-2. Therefore, \( \mathbf{W}_x \) satisfies Assumption 5 and is an adjustment set for unbiasedly estimating the PDE and PIE.


Based on Theorem 1, the counterfactual expressions for PDE (Eq.~\ref{doPDE}), PIE (Eq.~\ref{doPIE}), and STE can be simplified into the do-expressions. This leads to our second theoretical finding presented below.
\begin{align}
\label{eq:do_PDE}
\mathrm{PDE} = \mathbb{E}\left[ \mathbf{Y} \mid \mathrm{do}(\mathbf{W}_t=\mathbf{w}_t', \mathbf{W}_y=\mathbf{w}_y'), \mathbf{W}_x=\mathbf{w}_x) \right] \hspace{-22em} \nonumber \\
&\quad - \mathbb{E}\left[\mathbf{Y} \mid \mathrm{do}(\mathbf{W}_t=\mathbf{w}_t, \mathbf{W}_y=\mathbf{w}_y'), \mathbf{W}_x=\mathbf{w}_x) \right] \nonumber \\
&\quad \times P(\mathbf{W}_y = \mathbf{w}_y' \mid \mathrm{do}(\mathbf{W}_t = \mathbf{w}_t'), \mathbf{W}_x = \mathbf{w}_x) \nonumber \\
&\quad \times P(\mathbf{W}_x = \mathbf{w}_x)
\end{align}
\begin{align}
\label{eq:do_PIE}
\mathrm{PIE}= \mathbb{E}\left( \mathbf{Y} \mid \mathrm{do}(\mathbf{W}_t=\mathbf{w}_t, \mathbf{W}_y=\mathbf{w}_y), \mathbf{W}_x=\mathbf{w}_x \right) \hspace{-22em} \nonumber \\
&\quad \times \left[ P(\mathbf{W}_y=\mathbf{w}_y \mid \mathrm{do}(\mathbf{W}_t=\mathbf{w}_t'), \mathbf{W}_x=\mathbf{w}_x) \right. \nonumber \\
&\quad \left. - P(\mathbf{W}_y=\mathbf{w}_y \mid \mathrm{do}(\mathbf{W}_t=\mathbf{w}_t), \mathbf{W}_x=\mathbf{w}_x) \right]
\end{align}
\begin{equation}
\label{eq:do_STE}
\mathrm{STE}= E(\mathbf{Y} \mid \mathrm{do}(\mathbf{T}=\mathbf{t}')) - E(Y \mid \mathrm{do}(\mathbf{T}=\mathbf{t}))
\end{equation}

Based on Pearl's back-door adjustment formula and rules of do-calculus~\cite{pearl2009causality}, we have the theorem 1 to simplify do-expressions into probability expressions.

\paragraph{Theorem 2.} If we can derive \( p(\mathbf{T}, \mathbf{Y}, \mathbf{X}, \mathbf{W}_x, \mathbf{W}_y, \mathbf{W}_t \) from the causal DAG in Fig~\ref{fig:2}(b), then the PDF, PIE and STE can be identified from the data as follow:
\begin{align}
\label{eq:PDE'}
\mathrm{PDE} = \left[ E(\mathbf{Y} \mid  \mathbf{W}_t=\mathbf{w}_t', \mathbf{W}_y=\mathbf{w}_y', \mathbf{W}_x=\mathbf{w}_x) \right. \hspace{-18em} \nonumber \\
&\quad \left. - E(\mathbf{Y} \mid  \mathbf{W}_t=\mathbf{w}_t, \mathbf{W}_y=\mathbf{w}_y', \mathbf{W}_x=\mathbf{w}_x) \right] \nonumber \\
&\quad \times P(\mathbf{W}_y = \mathbf{w}_y' \mid  \mathbf{W}_t = \mathbf{w}_t', \mathbf{W}_x = \mathbf{w}_x) \nonumber \\
&\quad \times P(\mathbf{W}_x = \mathbf{w}_x)
\end{align}
\begin{align}
\label{eq:PIE'}
\mathrm{PIE}= E(\mathbf{Y} \mid \mathbf{W}_t=\mathbf{w}_t, \mathbf{W}_y=\mathbf{w}_y, \mathbf{W}_x=\mathbf{w}_x) \hspace{-18em} \nonumber \\
&\quad \times  \left[ P(\mathbf{W}_y=\mathbf{w}_y \mid \mathbf{W}_t=\mathbf{w}_t', \mathbf{W}_x=\mathbf{w}_x) \right. \nonumber \\
&\quad \left. - P(\mathbf{W}_y=\mathbf{w}_y \mid \mathbf{W}_t=\mathbf{w}_t, \mathbf{W}_x=\mathbf{w}_x) \right] 
\end{align}
\begin{align}
\label{eq:STE'}
\mathrm{STE}=  P(\mathbf{X}=\mathbf{x}) 
P(\mathbf{W}_x=\mathbf{w}_x) \hspace{-12em} \nonumber \\
&\quad \times \left[ E(\mathbf{Y} \mid \mathbf{T}=\mathbf{t}', \mathbf{X}=\mathbf{x}, \mathbf{W}_x=\mathbf{w}_x) \right. \nonumber \\
&\quad \left. -E(\mathbf{Y} \mid \mathbf{T}=\mathbf{t}, \mathbf{X}=\mathbf{x}, \mathbf{W}_x=\mathbf{w}_x) \right] 
\end{align}

\textit{Proof}: We prove that Eq~\ref{eq:do_PDE} to Eq~\ref{eq:do_STE} are identifiable, i.e., the do-operator can be converted to a do-free expression in Eq~\ref{eq:PDE'} to Eq~\ref{eq:STE'}, respectively. Our proposed causal DAG is shown in Fig~\ref{fig:2}(b). Based on rule 2 of do-calculus, we have \( P(\mathbf{Y}\ = \mathbf{y}\ \mid \text{do}(\mathbf{W}_t = \mathbf{w}_t', \mathbf{W}_y = \mathbf{w}_y'), \mathbf{W}_x=\mathbf{w}_x)) = P(\mathbf{Y}\ = \mathbf{y} \mid \text{do}(\mathbf{W}_t = \mathbf{w}_t'), \mathbf{W}_y = \mathbf{w}_y', \mathbf{W}_x=\mathbf{w}_x))\) since \( (\mathbf{Y}\ \perp\!\!\!\perp \mathbf{W}_y \mid \mathbf{W}_t, \mathbf{W}_x)_{G_{\overline{\mathbf{W}_t}\underline{\mathbf{W}_y}}} \),  where \(\overline{\mathbf{W}_t}\) represents the removal of all arrows pointing to \(\mathbf{W}_t\), and \(\underline{\mathbf{W}_y}\) represents the removal of all arrows emanating from \(\mathbf{W}_y\). 
\( P(\mathbf{Y}\ = \mathbf{y} \mid \text{do}(\mathbf{W}_t = \mathbf{w}_t'), \mathbf{W}_y = \mathbf{w}_y', \mathbf{W}_x=\mathbf{w}_x)) = P(\mathbf{Y}\ = \mathbf{y} \mid \mathbf{W}_t = \mathbf{w}_t', \mathbf{W}_y = \mathbf{w}_y', \mathbf{W}_x=\mathbf{w}_x))\) because  \((\mathbf{Y}\ \perp\!\!\!\perp \mathbf{W}_t \mid \mathbf{W}_y, W_x)_{G_{\underline{\mathbf{W}_t}}}\), where \(\underline{\mathbf{W}_t}\) represents the removal of all arrows emanating from \(\mathbf{W}_t\).
\(P(\mathbf{W}_y = \mathbf{w}_y' \mid \text{do}(\mathbf{W}_t = \mathbf{w}_t'), \mathbf{W}_x=\mathbf{w}_x)) = P(\mathbf{W}_y = \mathbf{w}_y' \mid \mathbf{W}_t = \mathbf{w}_t', \mathbf{W}_x=\mathbf{w}_x))\) because \((\mathbf{W}_y \perp\!\!\!\perp \mathbf{W}_t \mid\mathbf{W}_x)_{G_{\underline{\mathbf{W}_t}}}\).

Based on the back-door adjustment formula, \(P(\mathbf{Y} = \mathbf{y} \mid \text{do}(\mathbf{T} = \mathbf{t}'))\) is identifiable because all backdoor paths from \(\mathbf{T}\) to \(\mathbf{Y}\) are blocked by adjusting for \(\mathbf{X}\) and \(\mathbf{W}_x\). Specifically, \(\mathbf{T} \leftarrow \mathbf{X} \rightarrow \mathbf{Y}\) is blocked by \(\mathbf{X}\), \(\mathbf{T} \leftarrow \mathbf{W}_x \rightarrow \mathbf{W}_t \rightarrow \mathbf{Y}\) is blocked by \(\mathbf{W}_x\), \(\mathbf{T} \leftarrow \mathbf{W}_x \rightarrow \mathbf{W}_t \rightarrow \mathbf{W}_y \rightarrow \mathbf{Y}\) is blocked by \(\mathbf{W}_x\), \(\mathbf{T} \leftarrow \mathbf{W}_x \rightarrow \mathbf{W}_y \rightarrow \mathbf{Y}\) is blocked by \(\mathbf{W}_x\), and \(\mathbf{T} \leftarrow \mathbf{W}_x \rightarrow \mathbf{W}_y \leftarrow \mathbf{W}_t \rightarrow \mathbf{Y}\) is blocked by \(\mathbf{W}_x\). Hence, \( P(\mathbf{Y} = \mathbf{y} \mid \text{do}(\mathbf{T} = \mathbf{t}')) = P(\mathbf{Y} = \mathbf{y} \mid \mathbf{T} = \mathbf{t}') P(\mathbf{X} = \mathbf{x}) P(\mathbf{W}_x = \mathbf{w}_x) \). For details on the processes for calculating the PDE, PIE, and STE, please refer to the Appendix due to page limit.



\subsection{Implementation}
\begin{figure}[t]
	\centering
	\includegraphics[scale=0.128]{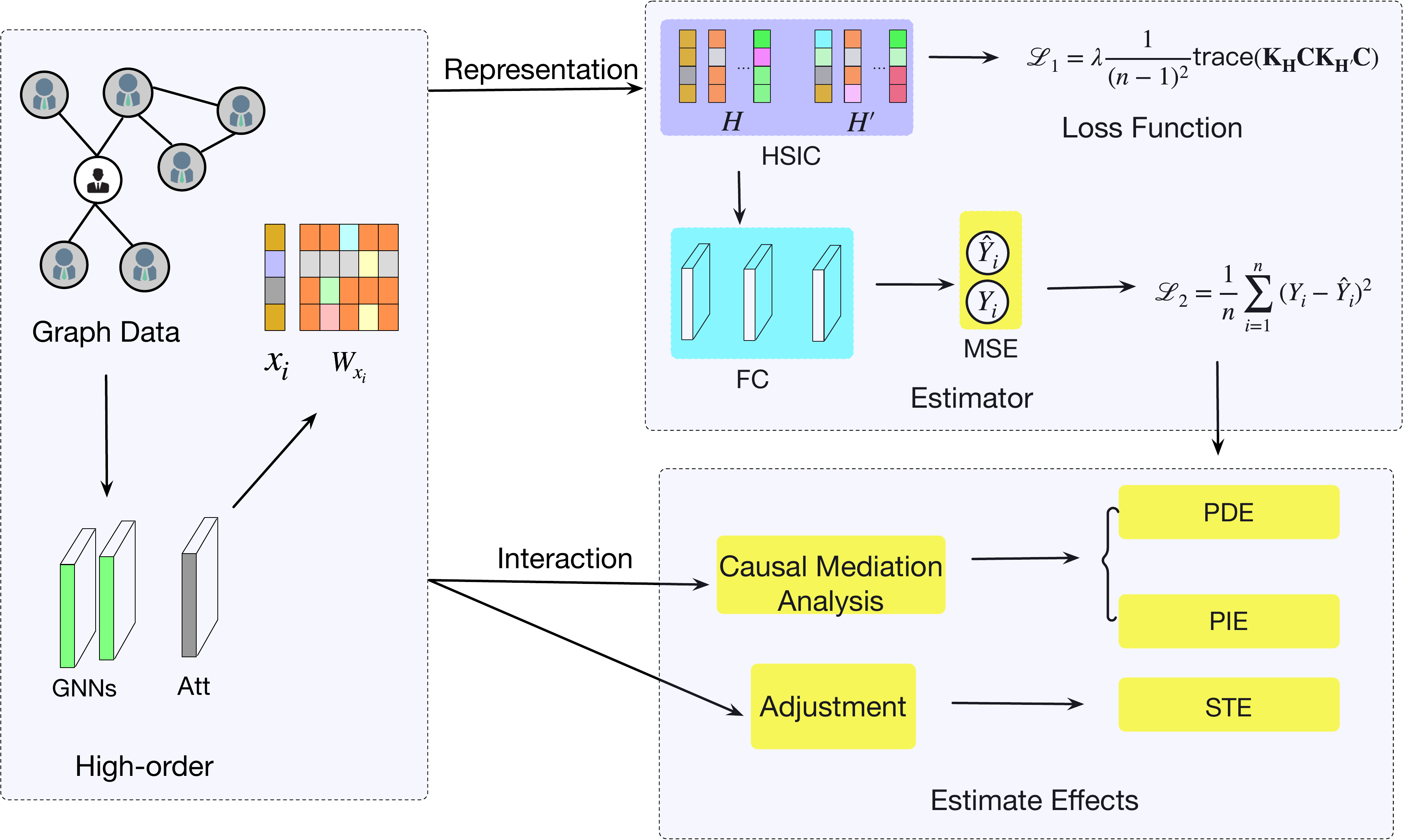}
	\caption{The workflow of our gDIS model for estimating PDE, PIE and STE within network data.}
	\label{fig:3}
\end{figure}

To accurately estimate PDE, PIE, and STE within networks, it is first necessary to establish Theorem 2,  proving that these effects can be effectively identified, and then derive Eq~\ref{eq:PDE'} to Eq~\ref{eq:STE'}. The implementation process involves three key steps: (1-1) \textit{Causal Mediation Analysis}: This step decomposes network peer effects into direct and indirect effects using causal mediation analysis~\cite{pearl2014interpretation}; (1-2) \textit{Back-Door Criterion Adjustment}: This step effectively identifies the STE using the back-door criterion, as introduced in the preliminaries. (2) \textit{GNNs}: Multi-layer GNNs with attention mechanisms are employed to capture the different influence of high-order neighbors. (3) \textit{HSIC Regularization}: HSIC regularization~\cite{ahmad2021regularized} is integrated to ensure independence between features and embeddings, maximizing the utilization of graph structures and enhancing model robustness. The workflow of our gDIS model is shown in Fig.~\ref{fig:3}, and it provides a robust framework for estimating group-level PDE, PIE, and STE.

\paragraph{Using attention weights to capture the varying influences of nodes.} A two-layer GNN with attention mechanisms dynamically weights the importance o neighboring nodes when updating each node's representation. By considering both direct and second-order neighbors, we calculate the attention coefficients \(\alpha_{ij}\) to indicate the relative importance of each neighbor as follows:
\begin{equation}
 \alpha_{ij} =  \frac{\exp(\phi({\mathbf{A}}^T [\mathbf{W} \mathbf{H}_i \, || \, \mathbf{W} \mathbf{H}_j]))}{\sum_{k \in \mathcal{N}(i) \cup \{i\}} \exp(\phi({\mathbf{A}}^T [\mathbf{W} \mathbf{H}_i \, || \, \mathbf{W} \mathbf{H}_k]))}
\label{eq:gat_update}
\end{equation}
where \(\alpha_{ij}\) indicates the importance of neighfbor \(j\) to node \(i\), and \(\phi(\cdot)\) is the \textit{LeakyReLU} activation function~\cite{xu2020reluplex}. \({\mathbf{A}}\) is a learnable weight vector used to calculate the unnormalized attention score between nodes, while \(\mathbf{W}\) is a learnable weight matrix that linearly transforms the feature vector \(\mathbf{H}\). \(\mathbf{H}_i\) and \(\mathbf{H}_j\) are the feature vectors of nodes \(i\) and \(j\), respectively, and \(||\) denotes vector concatenation. \(\mathcal{N}(i)\) represents the neighbor set of node \(i\).

The new feature representation of node \(i\) incorporates both its own information and that of its neighbors, resulting in a more comprehensive and accurate update.

\begin{equation}
H'_i = \sigma ( \sum_{j \in \mathcal{N}(i) \cup \{i\}} \alpha_{ij} \mathbf{W} \mathbf{H}_j)
\label{eq:update}
\end{equation}
where \(\sigma(\cdot)\) represents the \textit{ELU} activation function~\cite{ide2017improvement}.

\paragraph{HSIC Regularization.} HSIC~\cite{ahmad2021regularized} measures the dependence between two variables in a Reproducing Kernel Hilbert Space (RKHS)~\cite{berlinet2011reproducing}. Adding an HSIC regularization term to the loss function promotes independence between the original features and node embeddings, encouraging the embeddings to rely more on graph structure rather than node features, thereby reducing overfitting and improving robustness.

The feature matrix \( \mathbf{H} \) and the kernel matrix \( \mathbf{H}' \) using the Gaussian kernel~\cite{keerthi2003asymptotic} function are defined as follows:
\begin{equation}
(\mathbf{K}_{\mathbf{H}})_{il} = \exp \left( -\frac{\|\mathbf{H}_i - \mathbf{H}_l\|^2}{2\gamma^2} \right)
\label{eq:Kh}
\end{equation}
\begin{equation}
(\mathbf{K}_{\mathbf{H}'})_{il} = \exp \left( -\frac{\|\mathbf{H}'_i - \mathbf{H}'_l\|^2}{2\gamma^2} \right)
\label{eq:Kh'}
\end{equation}
where \( \|\mathbf{H}_i-\mathbf{H}_l\|^2\) is the squared euclidean distance~\cite{danielsson1980euclidean} between feature vectors \( \mathbf{H}_i \) and \( \mathbf{H}_l \). \( \gamma \) is the Gaussian kernel bandwidth~\cite{kakde2017peak}, set to the median of input feature distances. \( \mathbf{I} \) is the \( m \times m \) identity matrix, \( \mathbf{1} \) is an \( m \times 1 \) vector of ones, and \( m \) is the total number of samples. The centering matrix $\mathbf{C}$ is calculated as:
\begin{equation}
\mathbf{C} = \mathbf{I} - \frac{1}{n} \mathbf{1}\mathbf{1}^T
\label{eq:c}
\end{equation}
\paragraph{Objective Function.} We use mean squared error (MSE) loss~\cite{chicco2021coefficient} to measure the difference between actual and predicted values. Our final objective function is:
\begin{equation}
\begin{aligned}
\mathcal{L}_{\text{total}} & = \frac{1}{n} \sum_{i=1}^{n} (Y_i - \hat{Y}_i)^2  \\ & + \lambda \cdot \frac{1}{(n-1)^2} \text{trace}(\mathbf{K}_{\mathbf{H}} \mathbf{C} \mathbf{K}_{\mathbf{H}'} \mathbf{C})
\end{aligned}
\label{eq:loss}
\end{equation}
where \( \lambda \) is a hyperparameter set to 0.1. The trace term measures the dependence between the feature matrix \(\mathbf{H}\) and the embedding matrix \(\mathbf{H}'\).


\section{Experiments}

In this section, we evaluate the causal effect estimation capabilities of gDIS. We use standard evaluation metrics to compare the performance of gDIS against baseline models, validating its effectiveness.

\subsection{Experiments Setup}
\textbf{Datasets.} Each unit \(i\) has only one observed treatment \(T_i\), peer exposure \(Z_i\), and outcome \(Y_i\) (the factual outcome), making direct causal effect estimation challenging due to unobservable counterfactuals. Following~\cite{jiang2022estimating, chen2024doubly}, we use semi-synthetic datasets where the network structure (features and link) is real, but treatments and outcomes are simulated. The datasets include two real-world social networks: BlogCatalog and Flickr~\cite{li2015unsupervised}, with detailed descriptions provided in the Appendix.

Given the high-dimensional and sparse nature of the original features, we follow the approach in~\cite{jiang2022estimating, chen2024doubly} and apply Latent Dirichlet Allocation (LDA)~\cite{blei2003latent} to reduce the dimension to 10. The network is then divided into training, validation, and test sets using the METIS algorithm~\cite{karypis1998fast}. $T$ and $Y$ are simulated according to the procedure outlined in Fig.~\ref{fig:2}(a). Due to space limitations, further details are provided in the appendix.

\begin{table*}[t]
\centering
\scriptsize 
\begin{tabular}{@{}llccccccc@{}} 
\toprule
Data & Effects & CFR(+N) & ND(+N) & TARNET(+N) & NetEst & TNet & gDIS(-HSIC) & gDIS \\ \midrule
\multirow{4}{*}{\thead{BC \\ (within-sample)}} 
& peer & 0.3195\textsubscript{±0.0299} & 0.3488\textsubscript{±0.0249} & 0.2830\textsubscript{±0.0229} & 0.0890\textsubscript{±0.0252} & 0.1019\textsubscript{±0.0275} & 0.0750\textsubscript{±0.0224} & \textbf{0.0719\textsubscript{±0.0062}} \\
& peer direct & / & / & / & / & / & 0.0514\textsubscript{±0.0121} & \textbf{0.0487\textsubscript{±0.0092}} \\
& peer indirect & / & / & / & / & / & 0.0236\textsubscript{±0.0110} & \textbf{0.0232\textsubscript{±0.0056}} \\
& self-treatment & 0.5929\textsubscript{±0.0565} & 0.5571\textsubscript{±0.0855} & 0.4922\textsubscript{±0.0522} & 0.3322\textsubscript{±0.0461} & 0.2671\textsubscript{±0.1372} & 0.1973\textsubscript{±0.0800} & \textbf{0.1816\textsubscript{±0.0020}} \\ \cmidrule(lr){1-9}

\multirow{4}{*}{\thead{BC \\ (out-of-sample)}} 
& peer & 0.3184\textsubscript{±0.0259} & 0.3488\textsubscript{±0.0250} & 0.2898\textsubscript{±0.0263} & 0.1198\textsubscript{±0.0279} & 0.0984\textsubscript{±0.0248} & 0.0791\textsubscript{±0.0081} & \textbf{0.0628\textsubscript{±0.0090}} \\
& peer direct & / & / & / & / & / & 0.0422\textsubscript{±0.0091} & \textbf{0.0364\textsubscript{±0.0121}} \\
& peer indirect & / & / & / & / & / & 0.0369\textsubscript{±0.0052} & \textbf{0.0264\textsubscript{±0.0048}} \\
& self-treatment & 0.5913\textsubscript{±0.0559} & 0.5478\textsubscript{±0.0810} & 0.5611\textsubscript{±0.1431} & 0.2856\textsubscript{±0.0407} & 0.2653\textsubscript{±0.1253} & 0.1763\textsubscript{±0.0111} & \textbf{0.1738\textsubscript{±0.0126}} \\ \cmidrule(lr){1-9}

\multirow{4}{*}{\thead{Flickr \\ (within-sample)}} 
& peer & 0.2575\textsubscript{±0.0741} & 0.3011\textsubscript{±0.0651} & 0.2215\textsubscript{±0.0585} & 0.0917\textsubscript{±0.0072} & 0.1209\textsubscript{±0.0450} & 0.0771\textsubscript{±0.0177} & \textbf{0.0565\textsubscript{±0.0175}} \\
& peer direct & / & / & / & / & / & 0.0456\textsubscript{±0.0093} & \textbf{0.0290\textsubscript{±0.0109}} \\
& peer indirect & / & / & / & / & / & 0.0315\textsubscript{±0.0144} & \textbf{0.0275\textsubscript{±0.0095}} \\
& self-treatment & 0.2939\textsubscript{±0.0984} & 0.2896\textsubscript{±0.1077} & 0.2316\textsubscript{±0.0715} & \textbf{0.0772\textsubscript{±0.0118}} & 0.2875\textsubscript{±0.1912} & 0.1734\textsubscript{±0.0114} & 0.1619\textsubscript{±0.0165} \\ \cmidrule(lr){1-9}

\multirow{4}{*}{\thead{Flickr \\ (out-of-sample)}} 
& peer & 0.2646\textsubscript{±0.0732} & 0.3112\textsubscript{±0.0501} & 0.2237\textsubscript{±0.0588} & 0.0911\textsubscript{±0.0075} & 0.1216\textsubscript{±0.0730} & 0.0793\textsubscript{±0.0180} & \textbf{0.0550\textsubscript{±0.0196}} \\
& peer direct & / & / & / & / & / & 0.0458\textsubscript{±0.0081} & \textbf{0.0282\textsubscript{±0.0121}} \\
& peer indirect & / & / & / & / & / & 0.0335\textsubscript{±0.0148} & \textbf{0.0268\textsubscript{±0.0098}} \\
& self-treatment & 0.2973\textsubscript{±0.0918} & 0.2903\textsubscript{±0.1079} & 0.2255\textsubscript{±0.0599} & \textbf{0.0808\textsubscript{±0.0121}} & 0.2977\textsubscript{±0.1883} & 0.1882\textsubscript{±0.0115} & 0.1474\textsubscript{±0.0158} \\ \bottomrule
\end{tabular}
\caption{Causal effect estimation results. The $\epsilon_{PEHE}$ error is reported. The top-performing results are emphasized in bold. Note that ``/'' indicates the model is not applicable for this effect.}
\label{table1}
\end{table*}

\paragraph{Metrics.} We evaluate the algorithms using two metrics: MSE~\cite{chicco2021coefficient} and PEHE~\cite{grimmer2017estimating}. MSE, defined as \(\epsilon_{MSE} = \frac{1}{m} \sum_{i=1}^{m} (\hat{Y}_i - Y_i)^2\), measures counterfactual estimation accuracy. PEHE, defined as \(\epsilon_{PEHE} = \sqrt{\frac{1}{m} \sum_{i=1}^{m} \left[(\hat{Y}_i(t') - \hat{Y}_i(t)) - (Y_i(t') - Y_i(t))\right]^2}\), evaluates the precision in estimating heterogeneous effects. Here, \(\hat{Y}_i\) and \(Y_i\) denote estimated and ground truth outcomes, with lower values indicating better performance.

\paragraph{Baselines.} We compared our model with six baselines: (1) CFR~\cite{Shalit2017estimating}, the state-of-the-art for effect estimation on IID data using integral probability metrics (IPM) for distribution balancing; (2) TARNet~\cite{Shalit2017estimating}, a variant of CFR without IPM; (3) NetDeconf~\cite{guo2020learning}, an adaptation of CFR for network data using GNNs to encode confounders; CFR+(N), TARNet+(N), NetDeconf+(N), which add peer exposure to handle network interference; (4) NetEst~\cite{jiang2022estimating}, which applies adversarial learning to bridge graph ML and causal effect estimation; (5) TNet~\cite{chen2024doubly}, integrating target learning; (6) gDIS (-HSIC), our gDIS method without the HSIC module.
\paragraph{Implementation Details.} We use a 2-layer GNN with attention mechanisms and a 3-layer fully connected estimator, with hidden embeddings of size 32. The learning rate is set to 0.001 for all modules, and the Adam optimizer~\cite{zhang2018improved} is used. Each task is repeated five times, with average results and standard deviations reported. Experiments are conducted using PyTorch 1.7.0, Python 3.8, CUDA 11.0, on an RTX A4000 GPU (16GB), 12 vCPUs (Xeon Gold 5320 @ 2.20GHz), 32GB memory, and Windows OS.

\paragraph{Results.} Our gDIS model estimates PDE, PIE, total peer effects (sum of peer direct and indirect effects), and self-treatment effects at the group level. We evaluate ``within-sample'' estimates on training networks and ``out-of-sample'' estimates on testing networks. Table~\ref{table1} summarizes the $\epsilon_{PEHE}$ results on the BC and Flickr datasets. gDIS consistently outperforms baseline models in both settings, indicating that our objective function effectively reduces counterfactual prediction errors. Moreover, the HSIC module significantly improves performance on the test set, demonstrating greater effectiveness and robustness compared to other models. While baseline models estimate total peer effects, they fail to distinguish between direct and indirect effects.

\begin{figure}[t]
	\centering
	\includegraphics[scale=0.165]{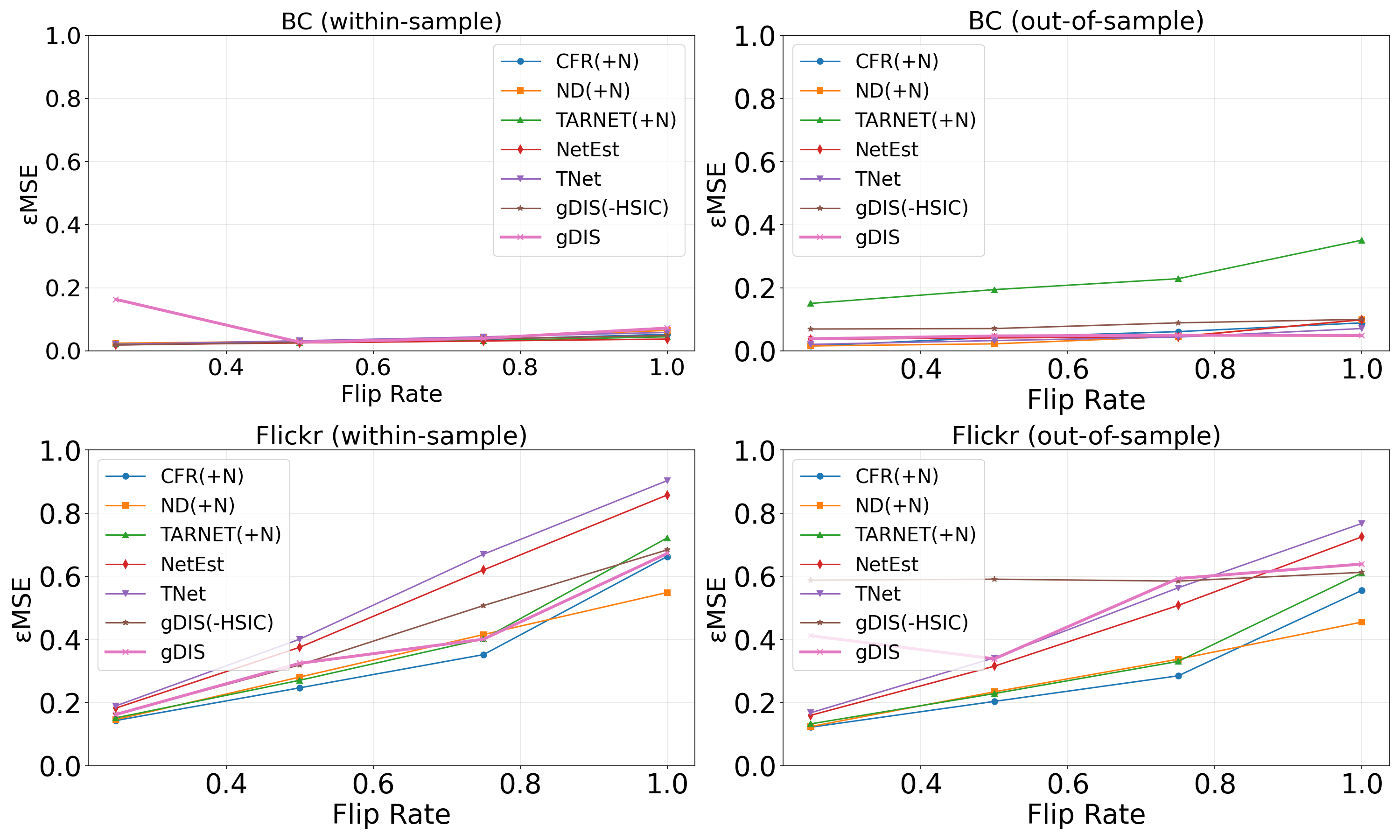}
	\caption{The results illustrate the relationship between the counterfactual estimation error ($\epsilon_{MSE}$) and the percentage of units with treatment flip.}
	\label{fig:4}
\end{figure}

In our counterfactual experiments, following~\cite{jiang2022estimating}, we simulated outcomes by varying the treatment flip rate (0.25, 0.5, 0.75, 1). As shown in Fig~\ref{fig:4}, higher flip rates generally lead to increased MSE, except in the case of the gDIS model under the ``Flickr out-of-sample'' setting, where the error remains stable, underscoring the HSIC module's role in improving generalization. Additionally, the Flickr dataset typically exhibits higher MSE compared to BC, likely due to its larger number of edge relationships.

\section{Related Work}
In this section, we review methods for estimating causal effects in network data that are related to our gDIS method.
\paragraph{Estimating Peer Effects in Network Data.} There are various methods for estimating peer effects in network data, where interference arises when an individual’s treatment influences the outcomes of connected individuals. For example, Forastiere et al.~\cite{forastiere2021identification} addressed this issue with a covariate adjustment method using a generalized propensity score (PS)~\cite{feng2012generalized} to balance both individual and neighborhood covariates. Jiang et al.~\cite{jiang2022estimating} introduced NetEst, a framework that employs graph neural networks (GNNs) to capture feature representations of both individual nodes and their first-order neighbors. Cai et al.~\cite{cai2023generalization} expanded on this by deriving generalization bounds and proposing a joint propensity score approach combined with representation learning via weighted regression. Ma et al.~\cite{ma2022learning} developed HyperSCI, which leverages hypergraph neural networks (HGNNs)~\cite{ma2022learning} to model interference using a multilayer perceptron (MLP)~\cite{popescu2009multilayer} and hypergraph convolution. However, none of these methods distinguish between direct and indirect peer effects.

\paragraph{Estimating Peer Direct and Indirect Effects in Network Data.} Another line of research focuses on estimating both direct and indirect peer effects in network data. VanderWeele et al.~\cite{vanderweele2012components} examined the effects of vaccination in small family units, such as the impact of a child’s vaccination on the mother. Shpitser et al.~\cite{Shpitser2017ModelingIV} developed a chain graph-based method to decompose peer effects into unit-specific components. Cai et al.~\cite{cai2021identification} evaluated contagion, susceptibility, and infectiousness effects in symmetric partnerships under infectious disease settings. Ogburn et al.~\cite{ogburn2024causal} analyzed obesity status in the Framingham Heart Study using longitudinal logistic regression, noting potential issues with dependence and model misspecification. However, these studies generally focus on scenarios with a propagation unit size of two, which may limit practical applicability. Ogburn et al.~\cite{ogburn2020causal} also used a log-linear model within the chain graph framework to evaluate the decisions of nine judges. However, the assumption of system equilibrium may not hold in dynamic environments, limiting the model's accuracy and applicability. Different from these reviewed works, our work focus on estimating peer direct and indirect effects at the group level. This more generalized setting allows us to capture the collective influence of groups, which is crucial for informing intervention strategies.
 
\section{Conclusion}
\textbf{Summary of Contributions.} In this work, we address the novel problem of estimating three types of causal effects in observational network data by proposing a new method, gDIS. Our approach differentiates between group-level PDE, PIE, and STE in observational network data. Through theoretical analysis, we establish identifiability conditions and provide corresponding proofs for the identifiability of these causal effects. To capture complex network interactions, gDIS employs a multi-layer GNN with attention mechanisms and incorporate HSIC to effectively control dependencies between node features. We validated the effectiveness and robustness of our gDIS on two semi-synthetic datasets, demonstrating strong performance even in complex network environments.

\textbf{Limitations \& Future Work.} While our framework is supported by theorems and the performance of gDIS has been demonstrated, there are limitations. Our approach relies on the network unconfoundedness assumption, which may be violated in practice, despite being common in the literature. In future work, we aim to relax these assumptions to expand the applicability of our method.

\bibliography{aaai25}

\end{document}